# A Modified Activation Function with Improved Run-Times For Neural Networks

**V.I.E Anireh**
Department of Computer Science
Rivers State University of Science and Technology
Port-Harcourt
Rivers State, Nigeria.
E-mail: viaemeka@gmail.com

**N.E. Osegi**
System Analytics Laboratories (SAL)
Sure-GP Ltd
Port-Harcourt
Rivers State, Nigeria.
E-mail: nd.osegi@sure-gp.com

**ABSTRACT**

In this paper we present a modified version of the Hyperbolic Tangent Activation Function as a learning unit generator for neural networks. The function uses an integer calibration constant as an approximation to the Euler number, e, based on a quadratic Real Number Formula (RNF) algorithm and an adaptive normalization constraint on the input activations to avoid the vanishing gradient. We demonstrate the effectiveness of the proposed modification using a hypothetical and real world dataset and show that lower run-times can be achieved by learning algorithms using this function leading to improved speed-ups and learning accuracies during training.

**Key Terms: Key Terms:** Adaptive Normalization, Hyperbolic Tangent Activation Function, Neural Networks, Real Number Formula, Vanishing Gradient Problem

1. **Introduction**

After many years of existence of Artificial Neural Network, the best way to implement it has remained one of issues the proponents are yet to come to agreement. From very simple to complex models, studies have shown that there are no clear cut guidelines for selecting any particular model.

Multilayer Perceptron's (MLPs) are Neural Networks (NNs) with one input layer, one hidden layer with a nonlinear transfer function, and one output layer with a linear transfer function Hornik et al (1989). They have the ability to approximate any function with a finite number of discontinuities. Nonlinear functions are normally used to model any natural state of affair and the problem of selecting their transfer function goes without any background theory. Specifically, issues relating to the development of MLP NNs based models and the selection of an appropriate nonlinear transfer functions, affects modeling, performance and consequently influence appreciation of results from such models.

The Back-propagation method which is a supervised training algorithm is by far the most commonly used method for training MLPs with nonlinear sigmoid function principally in their hidden layer. All sigmoid functions share a similar 'S' shape that is essentially linear in their center and nonlinear towards their bounds asymptotically. More precisely, training a network means minimizing the error of a cost function such as the sum of squares function and the computation of its derivative.

One of the common complaints about back-propagation is that it is slow. However, it has been used quite successfully on a wide range of problems more than any other algorithm. A lot of work has been done in search of faster methods including those documented by Reed et al (1998).

In this paper, we have modified the hyperbolic tangent function by providing an alternative replacement of the Euler function e, and also included an adaptive normalization routine. This produced a dramatic effect on the performance, stability and accuracy of the result obtained. Further, test on other transfer functions namely – exponential linear unit (ELU) and hyperbolic tangent function (HTAN) were conducted. Our observation showed that the modified HTAN function has made a strong statement in solving the problem of vanishing gradient (VGP) and speed up learning time.

2.  **Related Literature**

    **2.1 The Vanishing Gradient Problem**

The vanishing gradient problem (VGP) has been identified as a perennial issue in neural networks particularly with activation functions that are sigmoidal. Hochreiter (1991) and Schmidhuber et al(1997), provided a detailed account of this problem and proposed the Long-Short-Term-Memory (LSTM) network as a remedy to existing recurrent back-propagation network such as the Back-Propagation-Through-Time (BPTT) network of Williams and Zipser (1995), the Real-Time Recurrent Learning (RTRL) networks of Robinson and Fallside (1987) and the learning algorithm of Pearlmutter (1989).

More formally, the VGP problem describes a phenomenon that occurs when the input activation goes out of range i.e. blows-up due to exponential increase or decrease in the net weight product of the input-hidden chain. Thus, the gradient vanishes and learning becomes difficult and unstable.

This problem can lead to poor accuracies and slow learning, particularly in networks that use random weight perturbations. Here we propose "An adaptive activation with a modifiable exponential function" as a candidate solution.

    **2.2 Exponential-like Activation Functions**

Activation units employing exponential functions play a useful role in neural network learning systems. In this section we briefly describe three popular activation learning units or functions used in neural networks and later compare their performances on various metrics.

**The Logistic Sigmoid (Soft-Step)**

The Logistic Sigmoid or Soft-Step originally introduced by Verhulst (1845), is a popular activation function that has been used in the past years by many researchers working on feed-forward back-propagation neural networks and is very popular in hydrologic applications Yonaba et al (2010). Soft-Step is defined as:

$$f_a = \frac{1}{1+e^{-x}} \qquad (2.1)$$

.and its gradient is expressed as:

$$f_a = (1-f_a)*f_a \qquad (2.2)$$

This function has the desirable natural squashing property for diverse inputs - see Fig 2.1(a). However, due to the VGP problem, this function may not perform well in real world applications. Another problem with this type of function is its boundary-point limitation with a typical range of between 0 and 1. This may lead to slower response to network prediction and reduced accuracy. The VGP occurs because x is squashed exactly at 1. This phenomenon is depicted in Fig 2.1 (b).

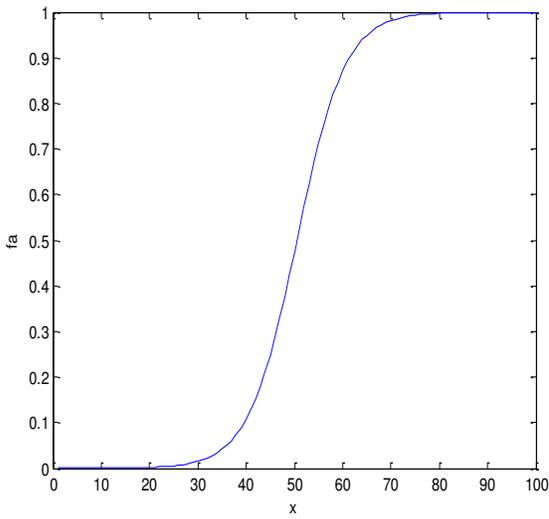 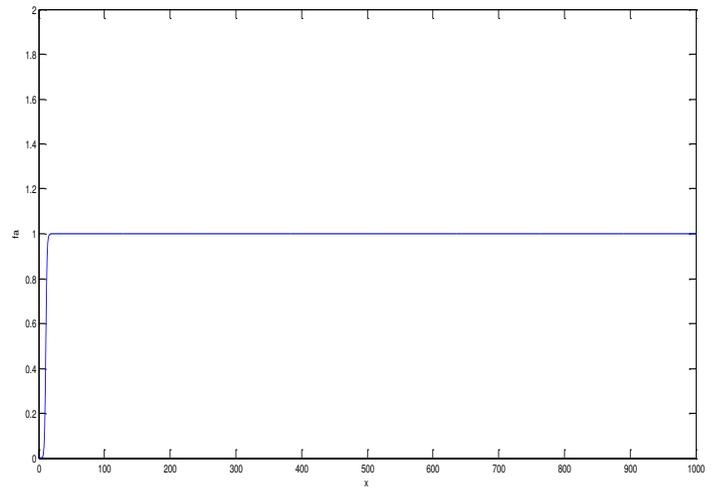

Fig 2.1 (a) Plot of Soft-Step for within-range values of x   Fig 2.1 (b) Plot of Soft-Step for exploding values of x

**Hyperbolic Tangent Function (HTAN)**

To overcome the limitations of the Soft-Step, HTAN was developed. HTAN is an activation function with a better range response than sigmoid leading to network speed-ups and more accurate predictions. The HTAN is defined as:

$$f_a = \frac{2}{1+e^{-2*x}} - 1 \qquad (2.3)$$

.and its gradient is expressed as:

$$f_a = 1 - f_a^2 \qquad (2.4)$$

A typical response curve for an HTAN is shown in Fig 2.2(a). However, just like the soft-step sigmoidal function, HTAN suffers from the VGP thus causing its activation response to stall at certain times and with reduced accuracy. The VGP case is depicted in Fig 2.2 (b) for exploding values of x.

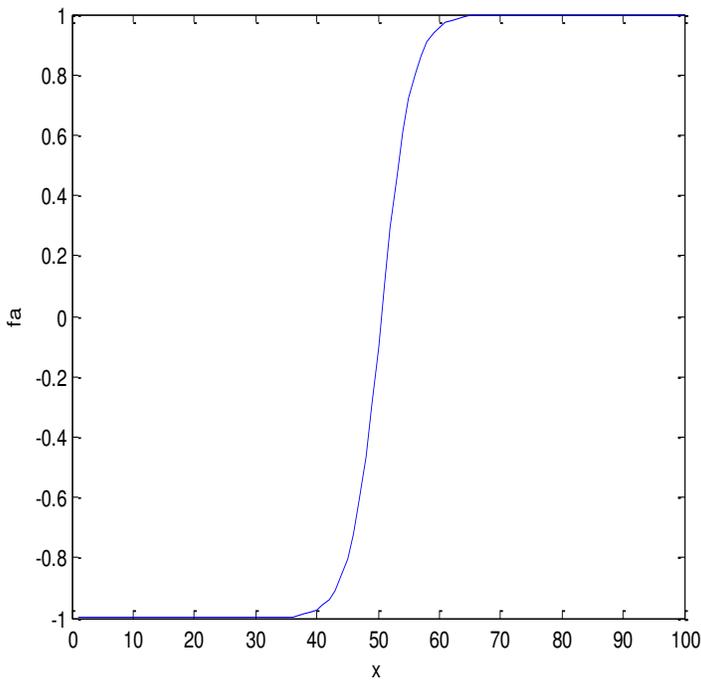

Fig 2.2 (a) Plot of HTAN for within-range values of x

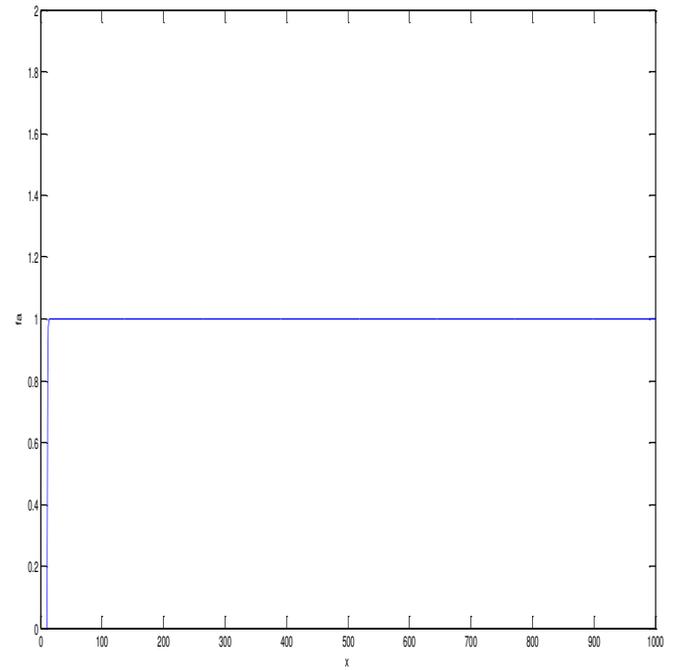

Fig 2.2 (b) Plot of HTAN for exploding values of x. Values of x is Between -10 and 1000

**Exponential Linear Unit (ELU)**

ELUs were introduced by Clevert et al (2015) in an attempt to solve the VGP. Specifically, an ELU is functionally defined as:

$$f_a = \begin{cases} x, & x > 0, \\ \alpha(e^x - 1) & x \leq 0 \end{cases} \tag{2.5}$$

.and its derivative (gradient) is computed as:

$$f_a' = \begin{cases} 1, & x > 0, \\ f_a + \alpha & x \leq 0 \end{cases} \tag{2.6}$$

The response of an ELU for values of x between -10 and +10 is as shown in Fig 2.3(a).

However, an ELU is still susceptible to VGP due to negative exploding weights (infinitesimal values), which may result in instabilities during training leading to network stalling. This is attributed to derivatives that cancel out when the values become -1. This situation is graphically depicted in Fig 2.3(b) for exploding values of x between -1000 and +1000. Stalling effect situation is due to the computation of these non-numeric activations and will be described in more detail in Section 5.

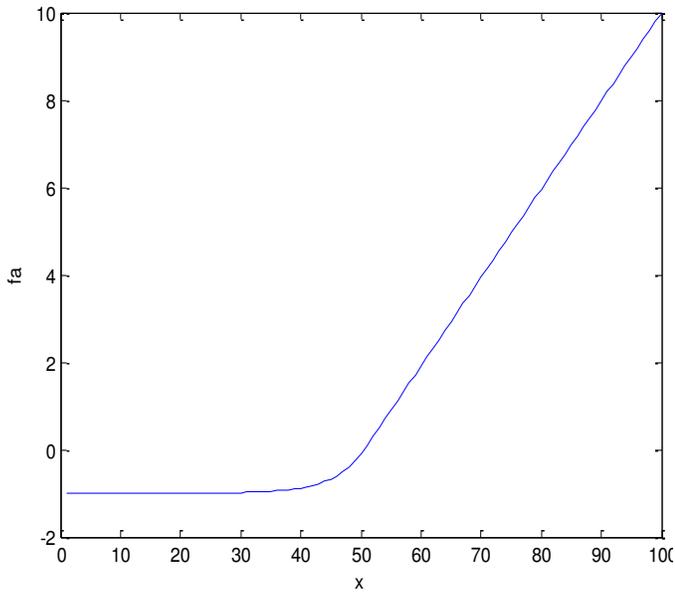 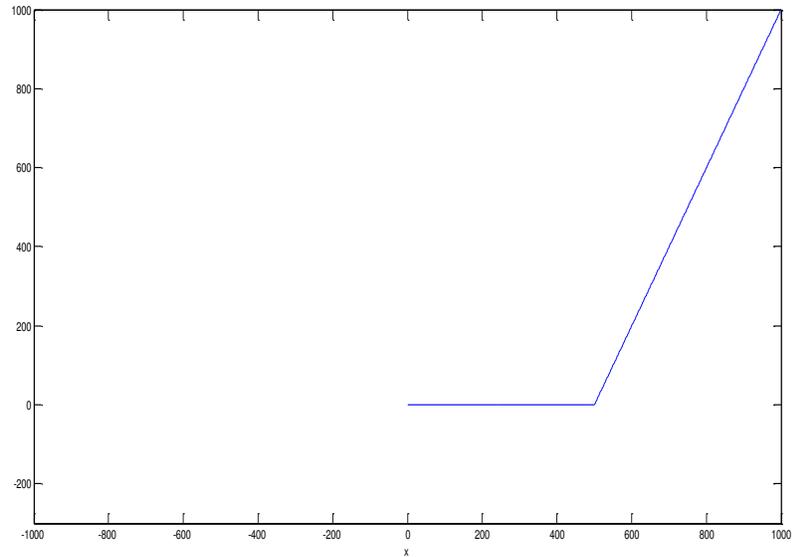

Fig 2.3 (a) Plot of ELU for within-range values of x

Fig 2.3 (b) Plot of ELU for exploding values of x

## 2.3 Machine or Algorithmic Representations of Exponential Function ($e^x$)

Several attempts have been made to improve the computation of e. While some approaches emphasize speed at the price of reduced accuracy, others prefer computation of large decimals of e with increased precision. In Schraudolph (1999), a machine based approach was proposed using a modified IEEE-754 floating point operation for the approximate computation of $e^x$ for neural activation functions. A revised version of this approach was proposed by Cawley (2000). However, due to the VGP, making $e^x$ fast is not sufficient to improve the overall neural network performance. Thus, several techniques and tricks have been recommended for eliminating or minimizing the vanishing gradient effect. This include but is not limited to batch normalization Hagan et al (1994), Lawrence et al (1997), using separate learning rates by Lecun et al (1998), the use of special gating networks by Hochreiter et al (1997) and more recently the attempts to improve the computation of $e^x$ through Single-Instruction Multiple Data (SIMD) architectures as in Malossi et al (2015). Thus, the computation of $e^x$ is still an active area of research.

### 3. Modified Hyperbolic Tangent Function with Adaptive Normalization

The hyperbolic tangent function (HTAN) is one of the many implicit-conditioned activation functions with a natural squashing operation for very large and small values beyond its range or continuous monotonicity for within range values. However, this function does not scale well for difficult learning tasks with potential explosive inputs leading to the VGP. The normalization trick described in Lecun et al (1998), is one attempt at avoiding this state but this may or may not be entirely useful, particularly for varying experimental models or datasets – for instance see Nayak et al (2014) and Nawi et al (2013). Thus, a better approach needs to be constructed from the basic principle. In this section we present an approach that eliminates the VGP exploding weights in the basic Hyperbolic Tangent Function which can lead to realistic results with reasonable accuracies.

**3.1 Modified Hyperbolic Tangent Function**

The Euler number (e for short) is a very vital mathematical function used widely in engineering and scientific research, as well as in industry. This constant is useful because of some interesting features such as good representation ability-replacing the structure of different kinds of functions or expressions, easy derivatives (when used as an exponential function). The latter is of interest to this presentation.

We approach the modification process from a different perspective focusing on the Real Number Formula (RNF) introduced in Osegi and Anireh (2016) with an adaptive constraint validated by the random weight method in Nguyen and Widrow (1990).

Using that idea in Osegi and Anireh (2016), we define the modified HTAN (MODHTAN) as:

$$\begin{cases} f_1 + f_2 + f_3, & f_1 = \left(\dfrac{k_o}{1+RNF^{-2*x_1}}\right) - 1, & x_1 = \dfrac{x}{(x+offset_1)} * (x \geq x_{cutoff}) \\ \\ & f_2 = \left(\dfrac{k_o}{1+RNF^{-2*x_2}}\right) - 1, & x_2 = \dfrac{x}{(x+offset_1)} * (x \leq -x_{cutoff}) \\ \\ & f_3 = \left(\dfrac{k_o}{1+RNF^{-2*x_3}}\right) - 1, & x_3 = \dfrac{x}{(x+offset_1)} * (x \geq -x_{cutoff} \vee x \leq +x_{cutoff}) \end{cases} \quad (2.7)$$

.where,

The *RNF* used here is described as:

$$RNF_o = \left(\dfrac{a-n}{a-(m+x)}\right)^a$$
$$\approx e^x \qquad (2.8)$$
$$for \quad a = 10^7, n = 1, and \, m = 1$$

.and $x_1$, $x_2$ are the constrained activations and $f_1$, $f_2$ the corresponding adaptive functions.

$RNF_o$ is cheaper to construct than $e^x$ and the derivative of HTAN can be used without any loss in precision. The parameters $offset_1$ and $x_{cutoff}$ represents the adaptive scaling (normalization) and threshold factors required for a typical squashing operation. The introduction of $offset_1$ provides immunity to the VGP defined by $x_{cutoff}$. Exploding values of weight are avoided as well so long as they fall within machine computable range. Typical values for $x_{cutoff}$ fall between the range of 10 and 100, but these values are not restrictive. A view of this function for within range and exploding values of x is shown in Figures 3.3(a) and 3.3(b) respectively.

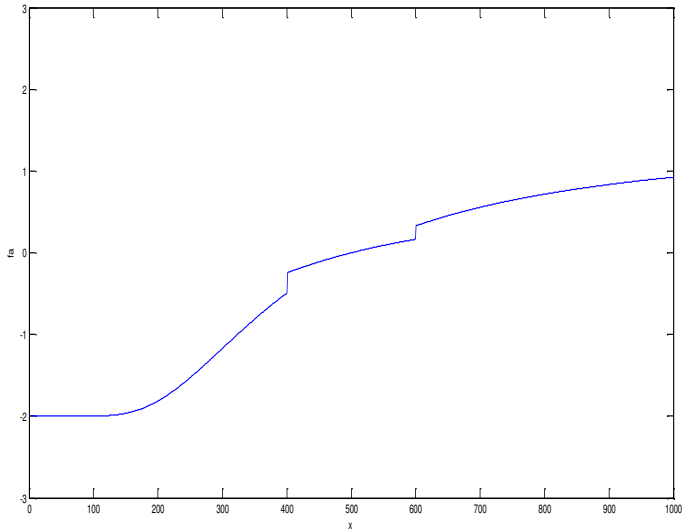 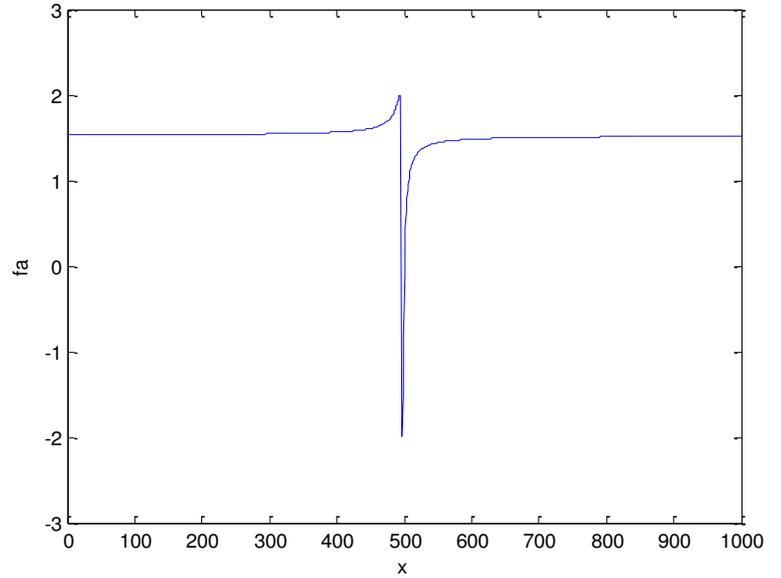

Fig 3.3 (a) Plot of MODHTAN for within-range values of x

Fig 3.3 (b) Plot of MODHTAN for exploding values of x

## 4. Experiments

**Training and Testing Environment:** Experiments were carried out on a system using Intel iCore3 processor, clock speed at 2.13GHz with 4GB Random Access Memory. The program was developed and run in Matlab ® 7.5 environment running on a Windows 7 Operating System.

**Test Instances:** Tests were performed on these three different activation functions namely – the Hyperbolic Tangent Function, Exponential Linear Unit, and our modified Hyperbolic Tangent Function. We used the back-propagation feed-forward neural network trained with the trainlm (Levenberg-Marquardt back propagation) function in Matlab and adaptive normalized constraint described in the previous section. The key parameters of the Neural Network employed are summarized in Table 3.1.

We have used a modified Euler number expression described in the previous section with negative exponents to derive an approximation to exponential function used in the sigmoidal and Hyperbolic Tangent Functions. Our test instances are of two-fold:

First, we use a synthetic dataset of 50,000 points linearly scaled between -1 and +1. For each simulation run, the time and corresponding error value is considered for each activation function. This process is carried out for 10 consecutive runs per activation function. Next, we use the approach in the first step using a benchmark dataset obtained from Blake et al (1998)

Table 3.1 Used Neural Network Parameters.

| Number of Hidden Neurons | Number of Epochs | Learning Parameter |
|---|---|---|
| 2 | 500 | Gradient descent with momentum |

5. **Results and Discussion**

**Hypothetical Dataset ($x^2 -2$):**

The run-times and error values of each function is as shown in Table 3.2, while Figures 3.4 and 3.5 shows the combined run-time and error plots compared for each activation respectively. The average values of each activation function are compared in Table 3.2. It can be seen that on the average HTAN performs best with least error while the MODHTAN performs best in speed.

Table 3.2 Run-time and prediction errors using the different activation functions for the hypothetical dataset.

| Run | HTAN | | ELU | | MODHTAN | |
|---|---|---|---|---|---|---|
| | **Run-time** | **Error** | **Run-time** | **Error** | **Run-time** | **Error** |
| 1 | 42.3939 | 0.0117 | 4.9017 | 0.0783 | 5.5257 | 0.1721 |
| 2 | 61.1079 | 0.0117 | 4.5973 | 0.0781 | 12.7868 | 0.0402 |
| 3 | 38.6627 | 0.0384 | 6.5751 | 0.0147 | 16.2045 | 0.0511 |
| 4 | 51.6584 | 0.0117 | 5.6723 | 0.0783 | 10.4317 | 0.0457 |
| 5 | 54.4989 | 0.0116 | 12.9322 | 0.0145 | 9.4488 | 0.0494 |
| 6 | 43.9991 | 0.0117 | 19.7028 | 0.0146 | 4.7662 | 0.1554 |
| 7 | 37.6886 | 0.0116 | 10.9102 | 0.0147 | 12.9735 | 0.0395 |
| 8 | 50.6950 | 0.0115 | 60.8943 | 0.0220 | 8.9041 | 0.0404 |
| 9 | 65.1129 | 0.0116 | 5.6627 | 0.0787 | 8.3687 | 0.0595 |
| 10 | 56.9262 | 0.0116 | 8.6928 | 0.0145 | 5.0611 | 0.1961 |
| **Average** | 50.2744 | **0.0143** | 14.0541 | 0.0408 | **9.4471** | 0.0849 |

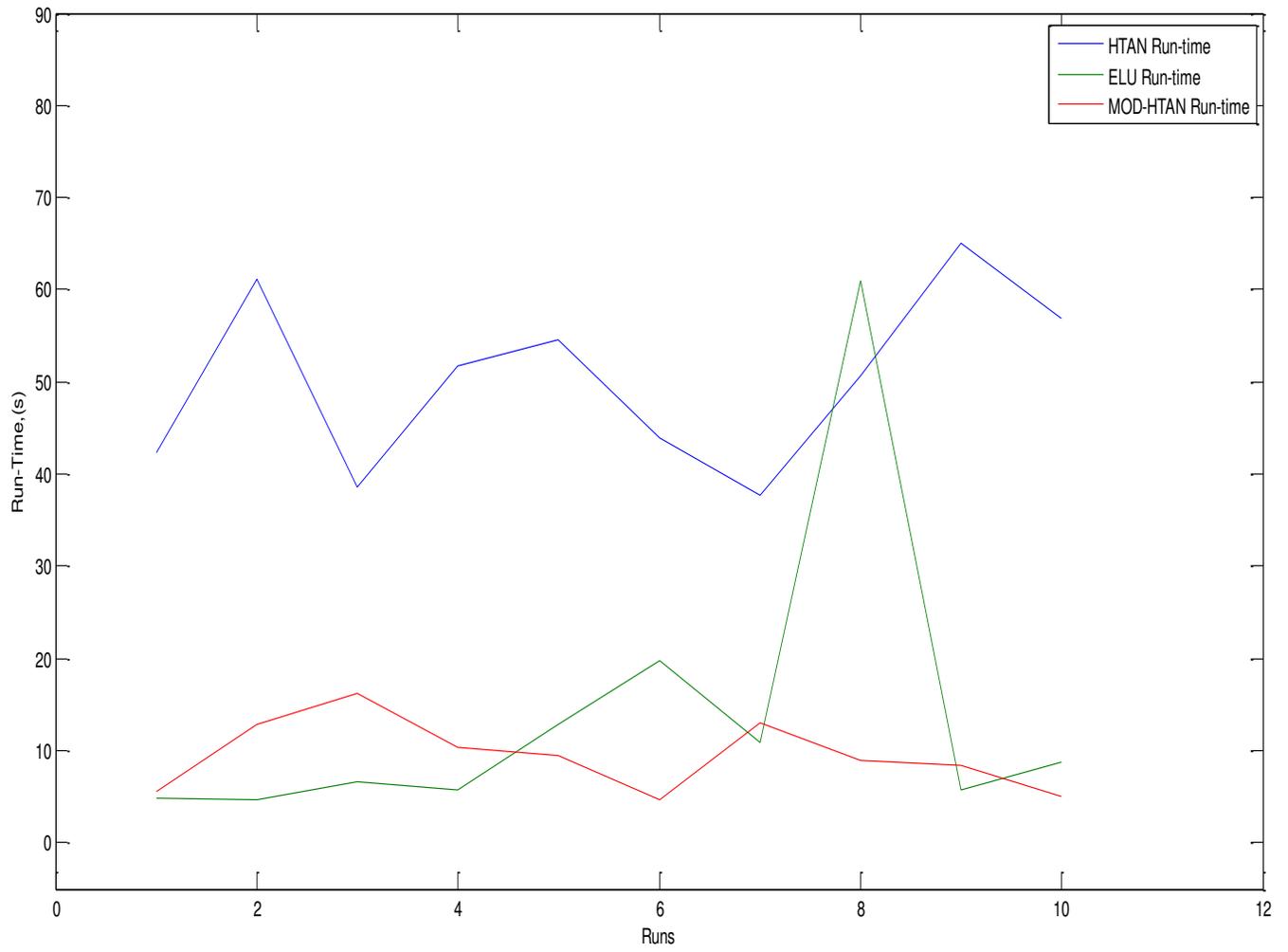

Fig 3.4 Combined run-time plots for each activation for hypothetical dataset (x^2 -2)

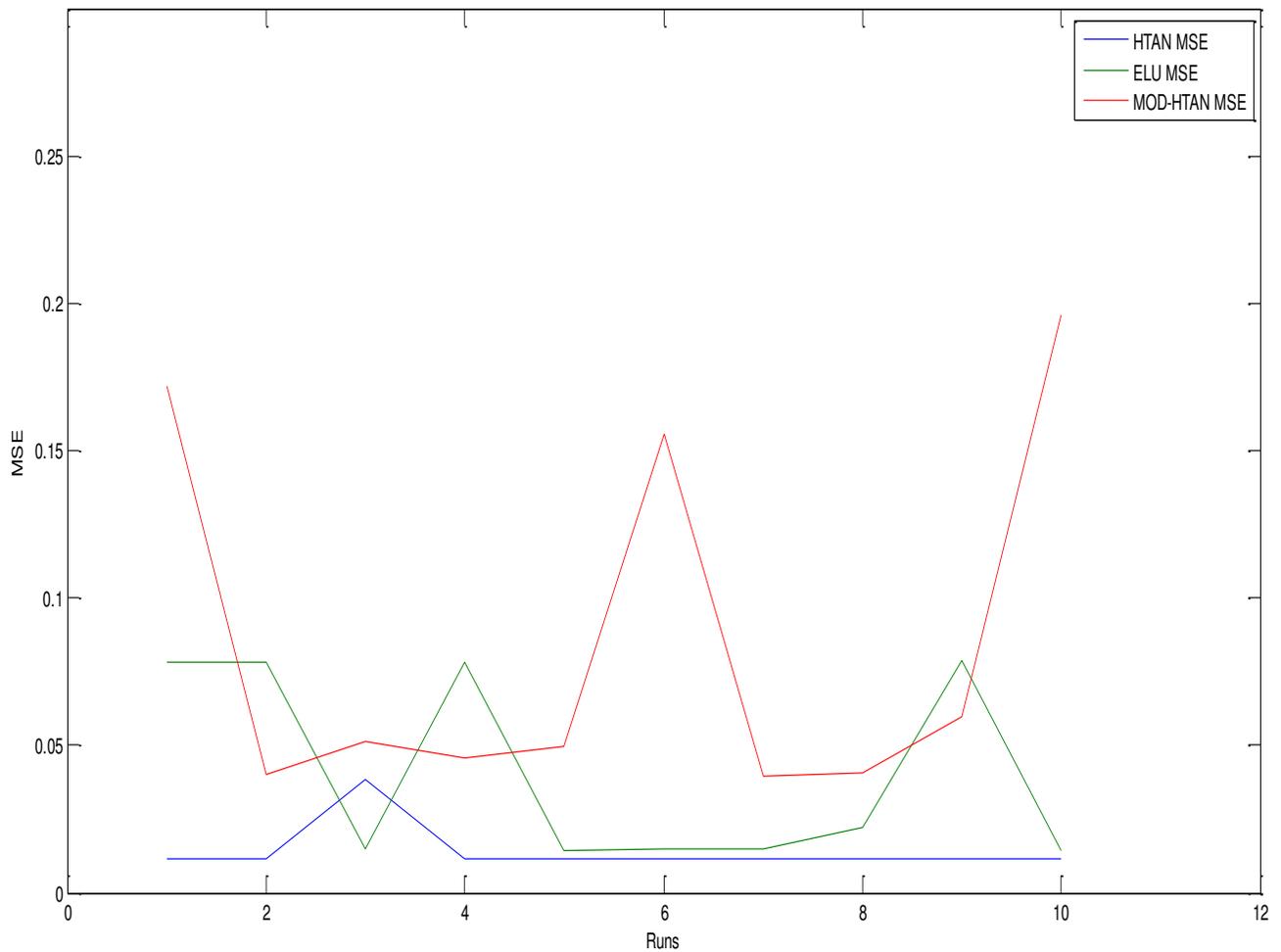

Fig 3.5 Combined Mean Squared Error (MSE) for each activation for hypothetical dataset ($x^2 -2$)

**Benchmark Dataset (Heart Dataset):**

The run-times and test classification accuracies of each function is as shown in Table 3.3, while Figures 3.6 and 3.7 shows the combined run-time and test classification plots compared for each activation respectively. The average values of each activation function are compared in Table 3.3 and 3.4. From these tables MODHTAN performed best in speed while ELU performs best in accuracy. It can also be seen from Fig3.7 that MODHTAN competes favorably well with the ELU in terms of accuracy.

Table 3.3 Run-time using the different activation functions for the Benchmark dataset.

| s/n | HTAN | ELU | MODHTAN |
|---|---|---|---|
| 1 | 5.4931 | 6.0055 | 4.9312 |
| 2 | 5.0896 | 4.0761 | 4.3900 |
| 3 | 4.4483 | 4.0761 | 3.3172 |
| 4 | 6.0440 | 4.5144 | 4.9215 |
| 5 | 5.6504 | 4.5111 | 5.8229 |
| 6 | 4.9325 | 5.0495 | 6.2570 |
| 7 | 4.9014 | 4.4460 | 3.5589 |
| 8 | 4.0921 | 5.3067 | 3.2816 |
| 9 | 5.2682 | 4.3292 | 4.0175 |
| 10 | 4.6140 | 4.6856 | 5.2798 |
| **AVERAGE** | 5.0534 | 4.7000 | **4.5778** |

Table 3.4 Classification-Test accuracies using the different activation functions for the Benchmark dataset.

| s/n | HTAN | ELU | MODHTAN |
|---|---|---|---|
| 1 | 75.92593 | 75.92593 | 70.37037 |
| 2 | 83.33333 | 75.92593 | 74.07407 |
| 3 | 79.62963 | 77.77778 | 74.07407 |
| 4 | 83.33333 | 88.88889 | 87.03704 |
| 5 | 61.11111 | 83.33333 | 75.92593 |
| 6 | 77.77778 | 83.33333 | 79.62963 |
| 7 | 83.33333 | 79.62963 | 81.48148 |
| 8 | 70.37037 | 88.88889 | 81.48148 |
| 9 | 75.92593 | 64.81481 | 81.48148 |
| 10 | 79.62963 | 83.33333 | 77.77778 |

| **AVERAGE** | 77.0370 | **80.1852** | 78.3333 |

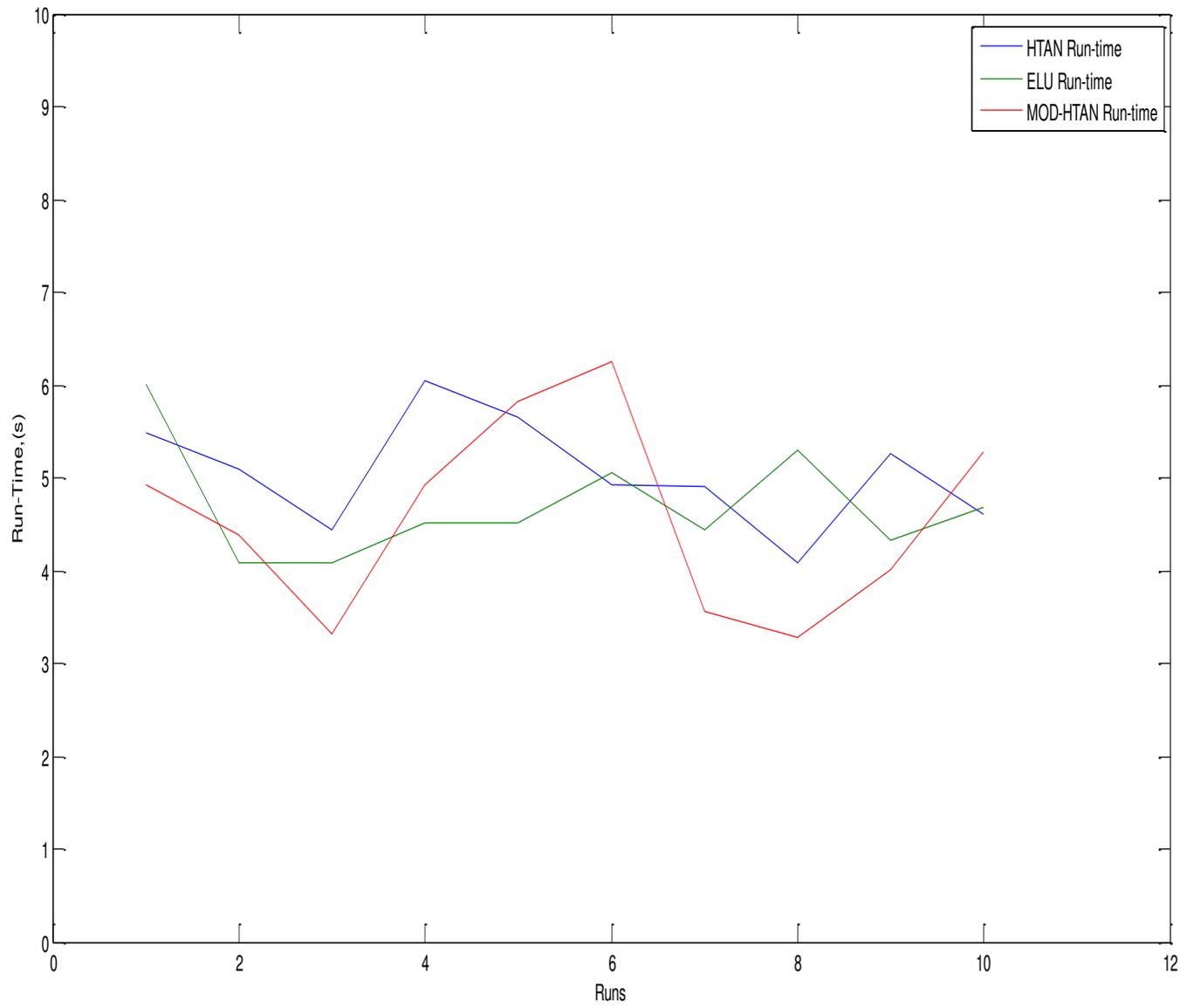

Fig 3.6 Combined run-time plots of each of the activation functions for the Heart dataset

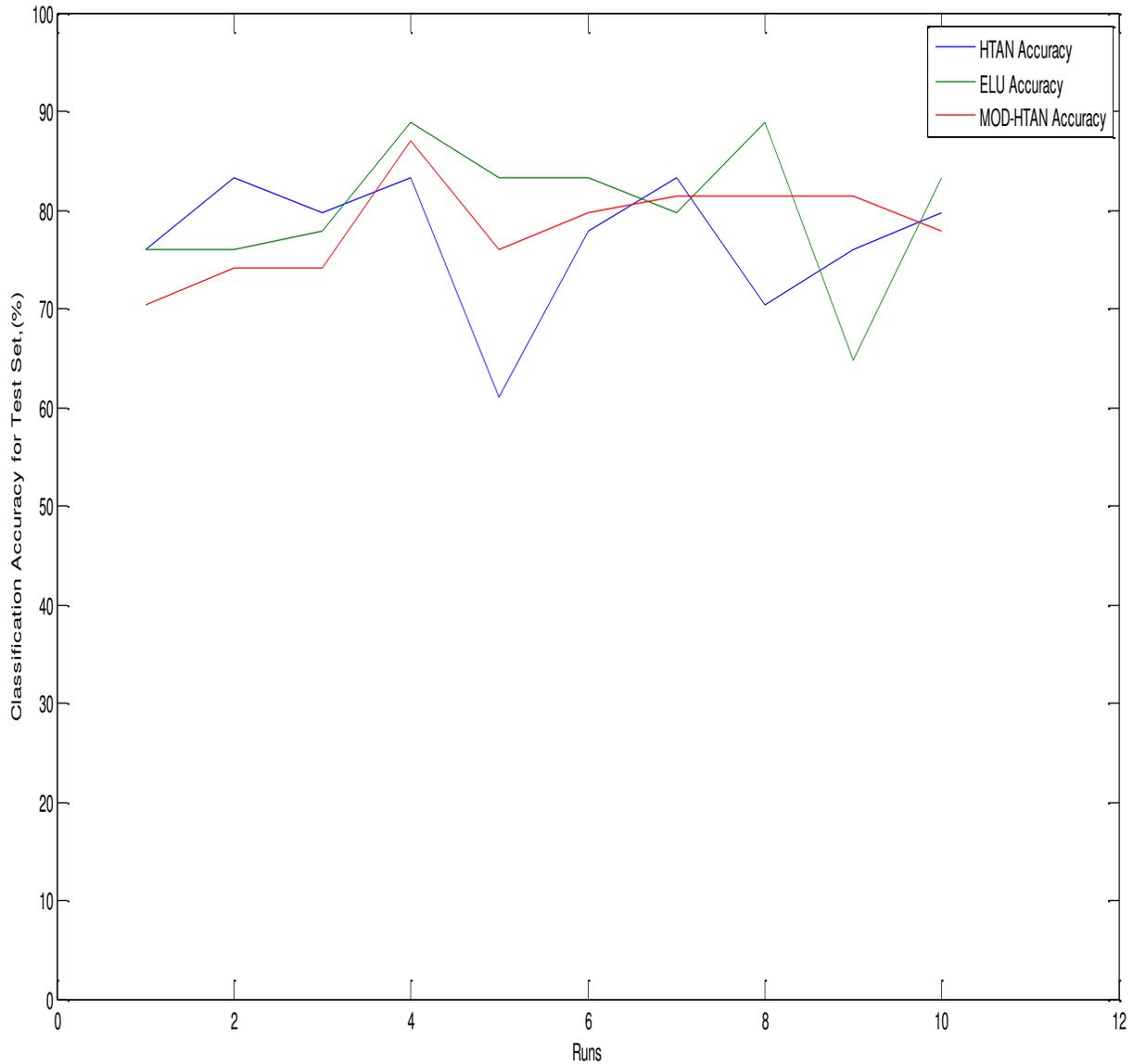

Fig 3.7 Combined Classification Accuracies of each of the activations function for Heart dataset

**Stalling Effect**

Due to VGP and exploding weights, long delays may be experienced prior to network training. Such undesirable points are peculiar characteristic feature of exponential functions that employ some sort of iterative algorithm such as the exponential function algorithm in Moler (2011).

**Conclusion**

From the foregoing, it is evident that many algorithms using the exponential function for computing its activations may lead to network learning failure in practical real world applications. This phenomenon can be avoided in the modified HTAN or activation functions using the RNF. Any previous use of RNF in network activation functions in any literature is unknown to the authors.

So far, this is a work in progress. We believe that these findings and effort towards improving the performance of neural networks transfer functions is worth paying attention.

It is also worthy to note, that this research work did not receive any specific grant from funding agencies in the public, commercial, or any not-for-profit sectors.